\documentclass[final]{cvpr}

\usepackage{times}
\usepackage{epsfig}
\usepackage{graphicx}
\usepackage{amsmath}
\usepackage{amssymb}
\usepackage{gensymb}

\usepackage{color}
\usepackage{subfigure}
\usepackage{threeparttable}
\usepackage{booktabs}
\usepackage{caption}
\usepackage[pagebackref=true,breaklinks=true,colorlinks,bookmarks=false]{hyperref}

\def\cvprPaperID{15} 
\def\confYear{CVPR 2021}

\begin{document}

\title{RSCA: Real-time Segmentation-based Context-Aware Scene Text Detection}

\author{Jiachen Li\textsuperscript{1}\thanks{Work is done during an internship at InnoPeak Technology},
Yuan Lin\textsuperscript{3},
Rongrong Liu\textsuperscript{3},
Chiu Man Ho\textsuperscript{3},
and Humphrey Shi\textsuperscript{1,2}\\
{\textsuperscript{1}UIUC, \textsuperscript{2}University of Oregon, \textsuperscript{3}OPPO US Research Center}\\
{\tt\small {jiachenl@illinois.edu} {\tt\small \{yuan.lin,rongrong.liu,chiuman\}@oppo.com}
{\tt\small {shihonghui3@gmail.com}}}}

\maketitle
\begin{abstract}
Segmentation-based scene text detection methods have been widely adopted for arbitrary-shaped text detection recently, since they make accurate pixel-level predictions on curved text instances and can facilitate real-time inference without time-consuming processing on anchors. However, current segmentation-based models are unable to learn the shapes of curved texts and often require complex label assignments or repeated feature aggregations for more accurate detection. In this paper, we propose RSCA: a Real-time Segmentation-based Context-Aware model for arbitrary-shaped scene text detection, which sets a strong baseline for scene text detection with two simple yet effective strategies: Local Context-Aware Upsampling and Dynamic Text-Spine Labeling, which model local spatial transformation and simplify label assignments separately. Based on these strategies, RSCA achieves state-of-the-art performance in both speed and accuracy, without complex label assignments or repeated feature aggregations. We conduct extensive experiments on multiple benchmarks to validate the effectiveness of our method. RSCA-640 reaches 83.9\% F-measure at 48.3 FPS on CTW1500 dataset.
\end{abstract}

\section{Introduction}
In recent years, scene text detection methods based on deep neural networks have been widely adopted in both academia and industry. Following the development of object detection and segmentation, representations for text instances in scene images rely on instance-level and pixel-level features that are extracted by deep convolutional neural networks. Pixel-level text representation learning, which are also known as segmentation-based methods, starts from EAST~\cite{zhou2017east} that removes anchors and makes multi-oriented text predictions directly from pixels to contours. Then, Textsnake~\cite{long2018textsnake} views text instances as sequences of ordered, overlapping disks and makes predictions on curved text. PSENet~\cite{Wang_2019_CVPR} encodes text spines on multiple scales and uses progressive scale expansion algorithm to reconstruct text instances. PAN~\cite{Wang_2019_ICCV} further enhances features with repeated feature fusion modules and DBnet~\cite{liao2020real} proposes differentiable binarization for boundaries of text instances. TextPerception~\cite{qiao2020text} also makes order-aware segmentation with labels on heads, contours and tails for arbitrary-shaped text detection. These segmentation-based methods set a general encoder-decoder prototype for arbitrary-shaped scene text detection and reach state-of-the-art performance on multiple curved scene text detection benchmarks.

\begin{figure}[tb]
\centering
\includegraphics[width=0.5\textwidth]{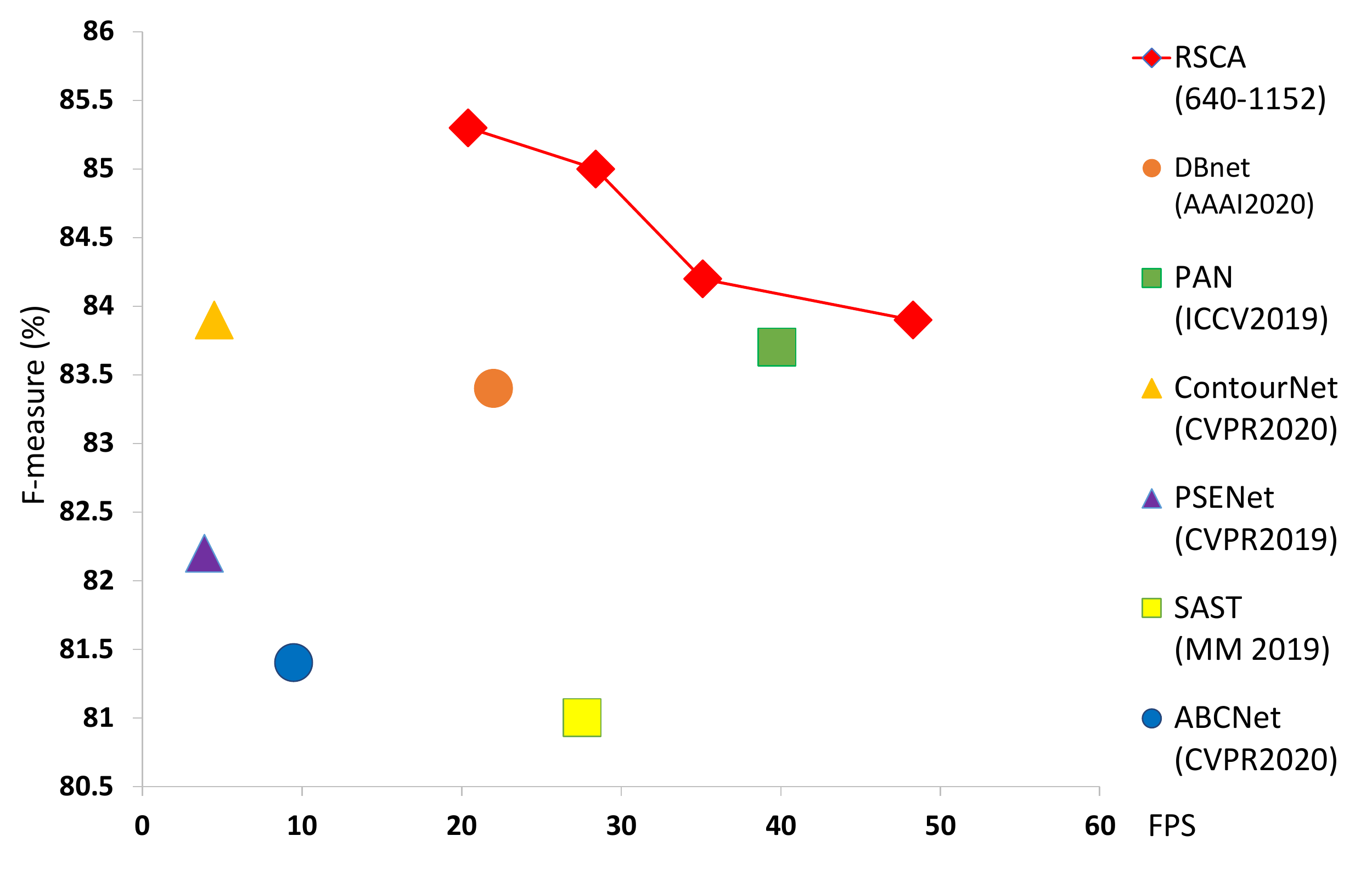}
\caption{Comparisons between RSCA and other state-of-the-art arbitrary-shaped scene text detection methods on CTW1500 benchmark.}
\label{Figure 1}

\end{figure}

However, among these segmentation-based methods for arbitrary-shaped scene text detection, there are two main problems. Firstly, it lacks modeling of curved shapes of text instances since common convolution and pooling layers only operate with fixed geometric structures, which are designed for predictions of regular bounding boxes. For curved scene text detection, since most instances are irregular polygons, models need ability to learn spatial transformation to reconstruct text polygons from segmentation maps, which is ignored by current state-of-the-art segmentation-based models. Secondly, label generation and assignment rules are complex and exhausting for arbitrary-shaped text instances. Different parts of text including heads, tails and boundaries are required to be generated manually and labeled as different classes. Text regions are shrunk with a fixed ratio as foregrounds for training process, which requires many hand-crafted parameters with grid search on different benchmarks to get state-of-the-art performances.

To tackle with these two problems, we propose two corresponding strategies: Local Context-Aware Upsampling (LCAU) and Dynamic Text-Spine Labeling (DTSL). For spatial transformation modeling, previous methods~\cite{jia2016dynamic}~\cite{fu2019dual} show that self-attention mechanism helps to model global pixel-to-pixel relation but is computationally expensive. Deformable convolution~\cite{dai2017deformable}~\cite{zhu2019deformable} predicts kernel offsets but it shares weights on entire feature maps and is sensitive to parameter initialization. We propose a local context-aware upsampling module that generates an attention weight matrix separately but computes locally on feature maps during upsampling process, which is light-weight compared to global self-attention layer while more effective and efficient according to our experiments. For simplifying label generation and assignment, we propose a simple dynamic text-spine labeling method, which simply shrinks text regions with a gradually increasing ratio during training process. This brings no additional computational burden but learns representations for text regions from easy samples to hard samples. These two strategies help us to build RSCA: a Real-time Segmentation-based Context-aware model for arbitrary-shaped scene text detection, which achieves state-of-the-art performances on multiple benchmarks with real-time inference speed.

To validate effectiveness of our method, we conduct extensive experiments on multiple benchmarks with our RSCA model. In Figure~\ref{Figure 1}, it shows that comparing with other state-of-the-art methods on arbitrary-shaped scene text detection, our RSCA achieves better performance with real-time inference on CTW1500 benchmark. More comparisons and experiments are presented in the following sections.

To summarize, our contributions are as follows:
\begin{itemize}
\item We analyze the problems of current segmentation-based models for arbitrary-shaped scene text detection: lack of spatial transformation modeling and complex label assignments.

\item We propose RSCA: a real-time segmentation-based context-aware model for arbitrary-shaped scene text detection with local context-aware upsampling and dynamic text-spine labeling, which models local spatial transformation and simplifies labels assignments with dynamically increasing text-spine labels separately.

\item We conduct extensive experiments on several benchmarks to validate effectiveness of our RSCA model which achieves state-of-the-art performances with real-time inference speed.

\end{itemize}

\section{Related Works}
In this section, we briefly review current scene text detection methods based on deep neural networks, including two main categories: anchor-based methods and segmentation-based methods.\\

\noindent \textbf{Anchor-based Methods:} Anchor-based methods mainly develop based on object detectors, which starts from Faster-RCNN~\cite{ren2015faster} that firstly introduces anchors as pre-defined boxes for accurate regression. Then, one-stage methods~\cite{zhang2019skynet}~\cite{li2021pseudo} employ anchors on feature pyramids and make predictions directly from anchors. Following their design, Textboxes~\cite{liao2017textboxes} changes anchor scales and follows SSD to detect text instances. Textboxes++~\cite{liao2018textboxes++} further employs quadrilateral regression for bounding boxes on multi-oriented text detection. RRD~\cite{Liao_2018_CVPR} decouples classification and regression branches for better multi-oriented text detection. To better handle arbitrary-shaped text detection, Mask TextSpotter~\cite{lyu2018mask} adds a segmentation branch for segmenting text instances from bounding boxes, which inherits from Mask RCNN~\cite{he2017mask}. ContourNet~\cite{wang2020contournet} proposes adopted RPN to generate more suitable anchors and segments contours for curved text detection. These anchor-based methods perform well on text detection with regular shapes, but lack robustness to arbitrary-shaped scene text detection, since both shape and aspect-ratio of the pre-defined anchors limit their potential for curved text detection.\\

\begin{figure*}[htb]
\centering
\includegraphics[width=1\textwidth]{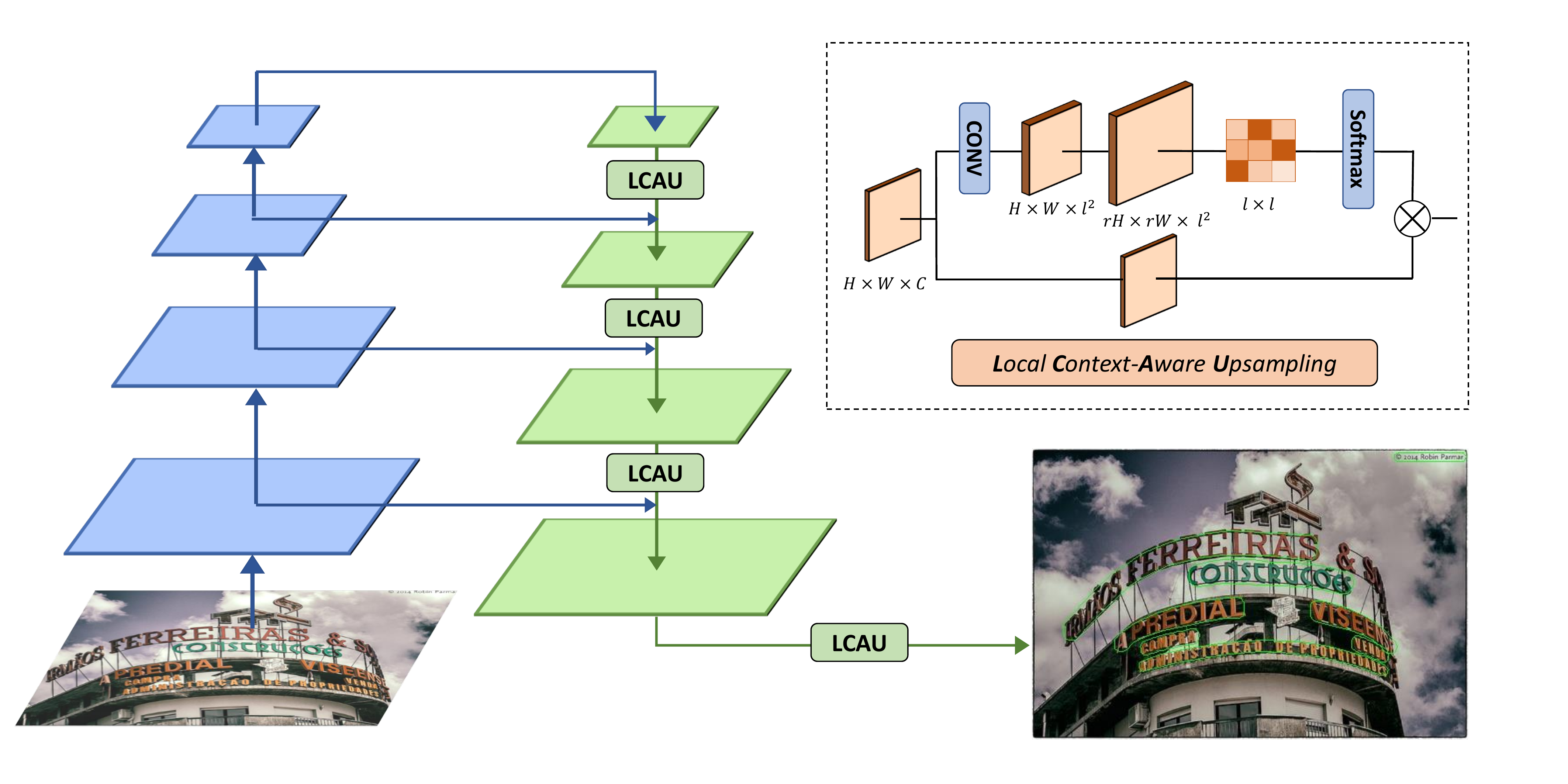}
\caption{RSCA model architecture with illustrations on Local Context-Aware Upsampling.}
\label{Figure 2}
\end{figure*}

\noindent \textbf{Segmentation-based Methods:} Segmentation-based methods mainly focus on pixel-level feature representation, which is suitable for arbitrary-shaped scene text detection since most text instances are curved. Following development of semantic segmentation, FCN~\cite{long2015fully}, U-Net~\cite{ronneberger2015u} and AlignSeg~\cite{huang2021alignseg} employ an encoder-decoder structure for pixel-level prediction. The encoder part is usually a deep feature extractor like ResNet~\cite{he2016deep} or VGG~\cite{simonyan2015deep} and the decoder part is usually feature upsampling by bilinear interpolation or deconvolution layer. EAST~\cite{zhou2017east} firstly removes anchors and make multi-oriented text instances prediction on pixels. Then,
more methods~\cite{long2018textsnake}~\cite{Wang_2019_CVPR}~\cite{Wang_2019_ICCV}~\cite{liao2020real}~\cite{qiao2020text}~\cite{xu2020rethinking} come out with focus on improving labeling accuracy and model assign. They push detection accuracy comparable to anchor-based methods on multiple scene text detection benchmarks. Among these segmentation-based methods, they set a baseline with FCN-like structure for pixel-level prediction and generate final detection results with grouping pixels to text instances. To ensure more accurate detection, they adopt repeated feature maps aggregation and complex label assignments, which could slow down the inference speed and limit their applications in broader scenarios due to specific requirements for labels.


\section{RSCA}
In this section, we mainly introduce our RSCA pipeline from model architecture to training and inference process, including our two effective strategies: local context-aware upsampling and dynamic text-spine labeling.

\subsection{Model Architecture}
We show our RSCA model architecture in Figure~\ref{Figure 2}. Specifically, we first adopt ResNet-50~\cite{ren2015faster} as our backbone for feature extraction with multiple levels of feature maps. It generates 5 stages of feature maps $C_1$, $C_2$, $C_3$, $C_4$, $C_5$ and the downsampling rate is $2^l$ for $C_l$ feature map. Then, following the feature pyramid design from FPN~\cite{lin2017feature}, we select $C_2$, $C_3$, $C_4$, $C_5$ for upsampling and feature aggregation. $C_1$ is not selected for reducing computational burden. During the feature aggregation stage, we use local context-aware upsamling to model pixel-to-pixel relation in a local range on each feature maps and concatenate augmented $C_2$, $C_3$, $C_4$, $C_5$ feature maps to $C_2$ scale with channel aggregation. Finally, we upsample aggregated $C_2$ to the original scale of the input image to predict text areas and reconstruct text instances.

\subsection{Local Context-Aware Upsampling}
Upsampling is a common operation in modern deep neural networks for computer vision tasks like object detection and semantic segmentation, since it promotes feature maps from low resolution to high resolution and from semantic level to pixel level. For arbitrary-shaped scene text detection, we propose local context-aware upsampling to model spatial transformation in a local range during upsampling, which improves detection accuracy especially on curved text regions.
\\

\begin{figure}[tb]
\centering
\includegraphics[width=0.45\textwidth]{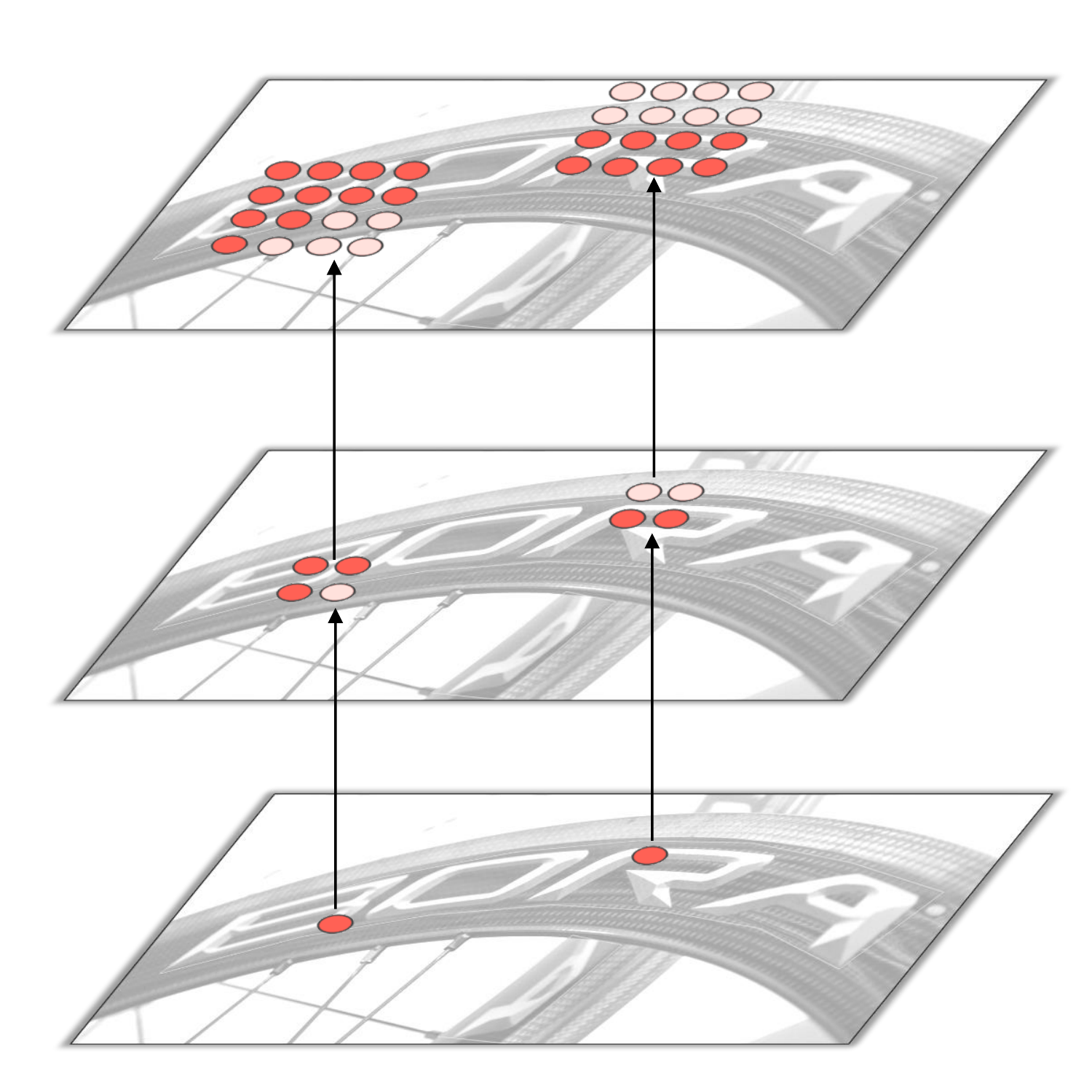}
\caption{Illustrations with local context-aware upsampling on original images. Shallow red point refers to non-text-region prediction and bright red point refers to text-region prediction.}
\label{Figure 3}
\end{figure}

\noindent \textbf{Upsampling Operators} For previous works in scene text detection, the most common upsampling operators are nearest neighbor and bilinear interpolations, which do not require any additional parameters. In Learning Deconvolution Network~\cite{noh2015learning}, it proposes deconvolution layer which is an inverse operator of convolution layer. It is learnable but applies the same kernel across the entire feature maps. In ESPCN~\cite{shi2016real}, it uses pixel shuffle as upsampling module which reshapes feature maps from the depth channel into width and height dimension. Our motivation is to model pixel-to-pixel relation in local range since arbitrary-shaped text are irregular and curved. Global self-attention~\cite{fu2019dual} is a decent solution but it introduces too much additional computational burden on the global spatial attention and channel attention matrices. Motivated by DCN~\cite{dai2017deformable}~\cite{zhu2019deformable}, CARAFE~\cite{wang2019carafe} and dynamic filter~\cite{jia2016dynamic}, we propose local context-aware upsampling, which models spatial transformation relation in a local range that does not bring too much computational burden. 
\\

\noindent \textbf{Local Context-Aware Upsampling} For a feature map $C \times H \times W$, after an upsampling operation, it becomes $C \times rH \times rW$ where $r$ is the upsampling rate. In Figure~\ref{Figure 2}, we display the entire feature processing flow of our local context-aware upsampling operation. On the weight matrix generation branch, we first apply a convolution layer with dimension $3 \times 3 \times C \times C^{'}$ where $C^{'} = l^2$, where $l$ is the receptive field of pixels on feature map $C \times H \times W$ and now the dimension is $C^{'} \times H \times W$. Then, we apply nearest neighbor upsampling operation with scaling rate $r$ and it becomes $l^2 \times rH \times rW$. On the depth channel, $l^2$ can be viewed as a weight matrix of coordinate $(x,y)$ on feature map $C \times rH \times rW$. Motivated by dynamic filter, we also add a softmax operation to normalize the weight matrix and apply a local context-aware matrix multiplication with original feature map $C \times H \times W$. Finally, the dimension of feature maps becomes $C \times rH \times rW$. Local context-aware upsampling can replace any upsampling operation by adding only a small computational burden. In Figure~\ref{Figure 3}, we make illustrations on original images, for classic nearest neighbor upsampling, predictions on text-region would activate fixed adjunct space on high-resolution feature maps, while local context-aware upsampling generates a local weight matrix that weakens activation of non-text context. More experiments with local context-aware upsampling are shown in ablation studies.

\subsection{Dynamic Text-Spine Labeling}
Label generation and assignment are important to scene text detection, especially on segmentation-based methods, since they dictate text regions for the model to learn from loss function. For arbitrary-shaped scene text detection, we propose a dynamic text-spine labeling method for label generation and assignment, which dynamically enlarges text-spine as labels during the training process. It provides more positive samples as training process goes from easy ones to hard ones and outperforms previous fixed text-spine labeling methods with cross-entropy loss.
\\

\noindent \textbf{Label Assignment for Segmentation-based Methods} For segmentation-based methods like PSENet~\cite{Wang_2019_CVPR} and DBnet~\cite{liao2020real}, they employ shrunk text instance masks as labels. For example, given a text instance $S$ with a group of vertices $\sum_i^n P_i$, we can compute its perimeter $L$ and area $A$ of the original polygon. Then, the shrunk offset~\cite{vatti1992generic} is 
$$
D = \frac{A}{L} (1 - r^2)
$$
where $r$ is the shrink ratio and set to be 0.4 in DBnet but different discrete values in PSENet. For text instance $S$, it is dilated with the shrunk offset $D$ to be $S_d$ and regions of $S_d$ are considered as text-spine labels for text. In loss function, a binary cross-entropy loss function with a ratio of positive to negative samples as 1 : 3 is employed:  
$$
L = \sum_i^n y_i \log x_i + (1 - y_i) \log (1-x_i)
$$
which is similar to salient object detection that views foregrounds of text as labels and predicts text regions during inference process.
\\

\begin{figure}[tb]
\centering
\includegraphics[width=0.48\textwidth]{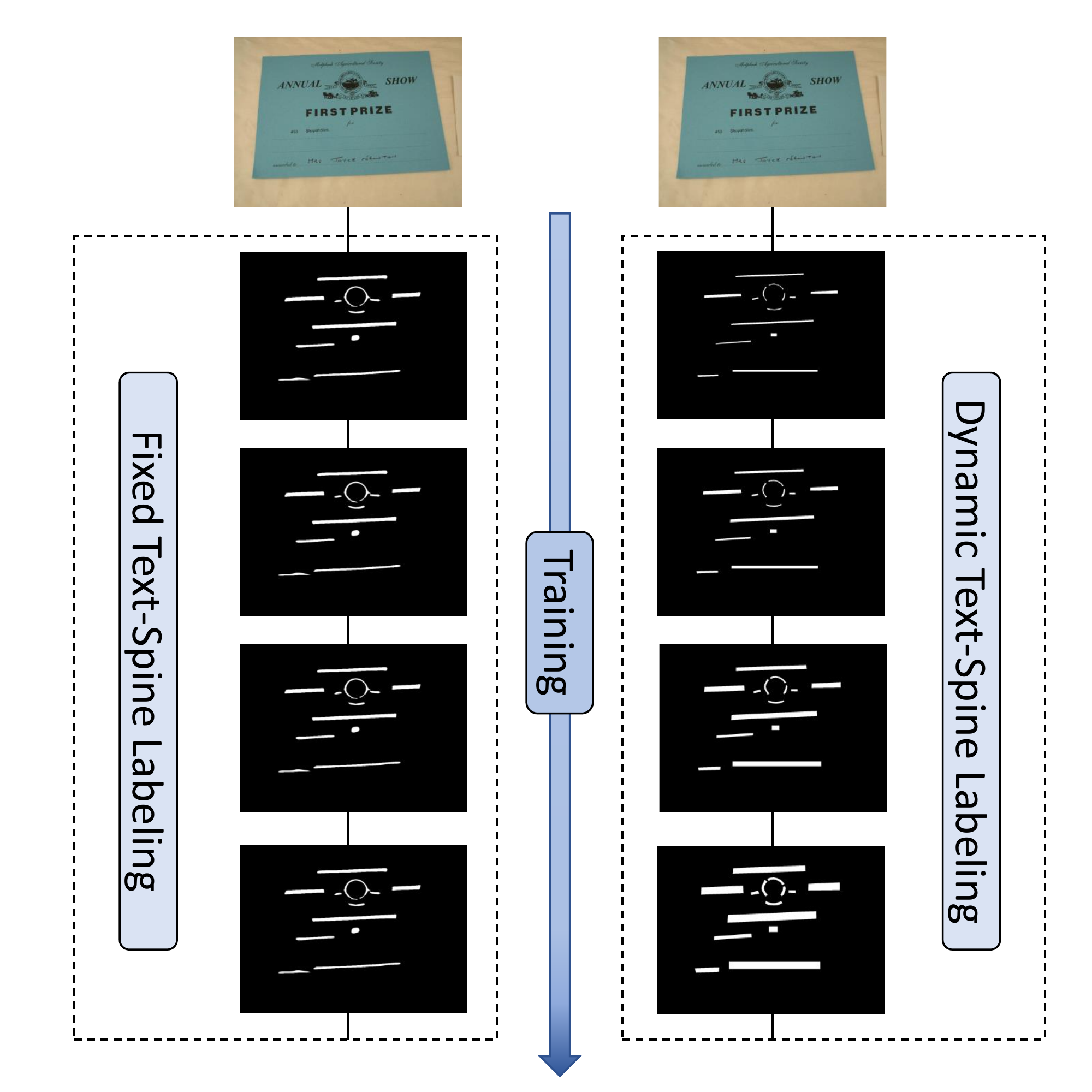}
\caption{Illustrations between dynamic and fixed text-spine labeling during training process.} 
\label{Figure 4}
\end{figure}

\noindent \textbf{Dynamic Text-Spine Labeling}~ Following the label generation from previous methods, there are two main drawbacks. Firstly, parameters of fixed shrink ratios vary with different datasets, which requires hand-crafted fine-tuning empirically. Secondly, splitting original text regions and label shrinking-boundary parts as negatives gives confusing signals to the scene text detector. To tackle with these two problems, we propose a dynamic text-spine labeling approach that enlarges text-spine labels during the training process. For the shrink ratio $r$, we set an initial value $r_a$ and a final value $r_b$. As training process goes on, we have
$$
r = r_a + (r_b - r_a) \times \frac{epoch}{max_{epoch}}
$$
The expansion coefficient $\beta = \frac{epoch}{max_{epoch}}$, which is motivated by the poly learning rate decay policy, decays exponentially as training continues. In Figure~\ref{Figure 4}, we make illustrations of both fixed and dynamic text-spine labeling methods during training process. Under this dynamic text-spine labelling approach, our scene text detection model could learn text regions from easy samples to hard samples, which introduces improvements compared with the setting of best fixed shrink ratio. For loss functions, after experiments with different ones with hard examples mining~\cite{lin2017focal}, we choose the basic binary cross-entropy loss with ratio of positive to negative samples as 1 : 3. More experiments with dynamic text-spine labeling and loss functions are shown in ablation studies.

\subsection{Inference}
During inference process, images are first resized to fixed size $(l,l)$, which is similar to the cropped samples during the training process. $l$ is set as 640 and 800 for our RSCA-640 and RSCA-800 model respectively. As shown in Figure~\ref{Figure 2}, the resized image is input to the RSCA model and it outputs a segmentation probability map with text-spine regions after feature extraction and local context-aware upsampling. To reconstruct text instances, we follow same steps used in DBnet~\cite{liao2020real}, which employs polygonal approximation algorithm to get independent text polygons, then dilates each individual text-spine according to its area $A_{ts}$, perimeter $L_{ts}$ and offset
$$
D_{ts} = \frac{A_{ts}}{L_{ts}} * d_{ts} 
$$
Here $d_{ts}$ is the dilation ratio. After dilating text instances, we reshape both the image and detection results into the original shape and get the final results.

\begin{table}[tb]
\centering
\begin{tabular}{c|c|c|c|c}
\toprule
Method    & Precision   &Recall  & F-measure  & FPS  \\ 
\midrule[1pt]
TextSnake~\cite{long2018textsnake}  &85.3  &67.9 &75.6 &- \\
NASK~\cite{cao2020all} &82.8 &78.3 &80.5 &12 \\
SAST~\cite{wang2019single} &85.3 &77.1 &81.0 &27.6 \\
CRAFT~\cite{baek2019character}  &86.0 & 81.1 &83.5 &- \\ 
DBnet-1024~\cite{liao2020real}  &86.9 &80.2 &83.4 &22 \\  
ABCNet~\cite{liu2020abcnet} &83.8 &79.1 &81.4 &9.5 \\  
PSENet~\cite{Wang_2019_CVPR} &84.8 &79.7 &82.2 &3.9 \\ 
ContourNet~\cite{wang2020contournet} &84.1 &\textbf{83.7} &83.9 &4.5 \\
PAN-640~\cite{Wang_2019_ICCV} &86.4 &81.2 &83.7 &39.8 \\ 
TextPerception~\cite{qiao2020text} &\textbf{88.8} &81.8 &\textbf{85.2} &- \\ \hline
RSCA-640 &87.2 &80.8 &83.9 &\textbf{48.3} \\  
RSCA-800  &87.2 &82.9 &85.0 &28.4 \\

\bottomrule
\end{tabular}

\caption{Detection results on CTW1500 dataset. All results are collected from CTW1500 leaderboard. The number with dash is the height of input images and \textbf{bold} indicates best results. }
\label{Table 1}
\end{table}

\begin{table}[tb]
\centering
\begin{tabular}{c|c|c|c|c}
\toprule
Method    & Precision   &Recall  & F-measure  & FPS  \\ 
\midrule[1pt]
CRAFT~\cite{baek2019character}  &87.6 & 79.9 &83.6 &- \\ 
DBnet-800~\cite{liao2020real}  &87.1 &82.5 &84.7 &32 \\  
ABCNet~\cite{liu2020abcnet} &85.4 &80.1 &82.7 &9.5 \\  
PSENet~\cite{Wang_2019_CVPR} &84.8 &79.7 &82.2 &3.9 \\ 
PAN-640~\cite{Wang_2019_ICCV} &89.3 &81.0 &85.0 &39.6 \\ 
ContourNet~\cite{wang2020contournet} &86.9 &83.9 &85.4 &3.8 \\
Boundary~\cite{wang2020all} &88.9 &85.0 &87.0 &- \\
E2E~\cite{qin2019towards} &87.8 &85.0 &86.4 &- \\
TextFuseNet~\cite{ye2020textfusenet} &89.0	&85.3 &87.1 &3.3 \\
CRAFTS~\cite{baek2020character} &\textbf{89.5}	&\textbf{85.4} &\textbf{87.4} &- \\\hline
RSCA-640 &86.9 &78.5 &82.5 &\textbf{40.3}\\  
RSCA-800 &86.6 &83.3 &85.0 &30.4 \\

\bottomrule
\end{tabular}

\caption{Detection results on Total-Text dataset. All results are collected from Total-Text leaderboard. Hyper-parameters of RSCA are adopted directly from CTW1500.}
\label{Table 2}
\end{table}

\section{Experiments}
In this section, we firstly introduce datasets and benchmarks that we use to validate the effectiveness of our method. Then, we show our experimental details including most hyper-parameters and hardware configurations. Furthermore, we compare our methods with other state-of-the-arts and present ablation study mainly on local context-aware upsampling and dynamic text-spine labeling.

\subsection{Datasets}
\noindent \textbf{CTW1500:} CTW1500~\cite{yuliang2017detecting} is also known as SCUT-CTW1500, which is a text-line based arbitrary-shaped text dataset with both English and Chinese instances. It contains 1000 training images and 500 testing images. Text instances are labeled with 14 points as polygons that can be described as arbitrary-shaped curve text.
\\
\noindent \textbf{Total-Text:} Similar to CTW1500, Total-Text~\cite{ch2017total} is an arbitrary-shaped text dataset but with word-level label. It contains 1255 training images and 300 testing images. Word instances are labeled with 10 vertices as polygons for curved text detection.
\\
\noindent \textbf{ICDAR 2015:} ICDAR 2015~\cite{karatzas2015icdar} is commonly used for multi-oriented text detection. It contains 1000 training images and 500 testing images. All text regions are annotated by 4 vertices of quadrangle.
\\
\noindent \textbf{MSRA-TD500:} MSRA-TD500~\cite{yao2012detecting} is a multi-language dataset that includes 300 images for training and 200 images for testing with text-line level labels. Following previous methods, we also include HUST-TR400~\cite{yao2014unified} in the training set with 400 images.
\\
\noindent \textbf{SynthText:} SynthText~\cite{gupta2016synthetic} is a synthetic dataset with 800000 images, which are synthesized on scene text with 8000 background images. SynthText is mainly used for pre-training our model.

\begin{table}[tb]
\centering
\begin{tabular}{c|c|c|c|c}
\toprule
Method    & Precision   &Recall  & F-measure  & FPS  \\ 
\midrule[1pt]
EAST~\cite{zhou2017east}  &87.3 &67.4 &76.1 &13.2 \\
TextSnake~\cite{long2018textsnake} &83.2 &73.9 &78.3 &1.1 \\ 
RRD~\cite{Liao_2018_CVPR}  &87.0 &73.0 &79.0 &10.0 \\
CRAFT~\cite{baek2019character}  &88.2 &78.2 &82.9 &8.6 \\ 
DBnet-736~\cite{liao2020real}  &91.5 &79.2 &84.9 &32 \\  
PAN-640~\cite{Wang_2019_ICCV} &84.4 &83.8 &84.1 &30.2 \\ 
ContourNet~\cite{wang2020contournet} &86.9 &83.9 &85.4 &3.8 \\ \hline
RSCA-640 &\textbf{92.8} &80.1 &86.0 &\textbf{52.5} \\  
RSCA-800 &91.5 &\textbf{85.6} &\textbf{88.4} &28.9 \\

\bottomrule
\end{tabular}
\caption{Detection results on MSRA-TD500 dataset. All results are collected from original papers. Hyper-parameters of RSCA are adopted directly from CTW1500.}
\label{Table 3}
\end{table}

\begin{table}[tb]
\centering
\begin{tabular}{c|c|c|c|c}
\toprule
Method    & Precision   &Recall  & F-measure  & FPS  \\ 
\midrule[1pt]
EAST~\cite{zhou2017east}  &83.6 &73.5 &78.2 &13.2 \\
TextSnake~\cite{long2018textsnake} &84.9 &80.4 &82.6 &1.1 \\ 
CRAFT~\cite{baek2019character}  &89.8 & 84.3 &86.9 &- \\
DBnet-1152~\cite{liao2020real}  &91.8 &83.2 &87.3 &12 \\  
RRD~\cite{Liao_2018_CVPR}  &85.6 &79.0 &82.2 &6.5 \\
PSENet~\cite{Wang_2019_CVPR} &86.9 &84.5 &85.7 &1.6 \\ 
PAN-640~\cite{Wang_2019_ICCV} &84.0 &81.9 &82.9 &26.1 \\ 
ContourNet~\cite{wang2020contournet} &87.6 &86.1 &86.9 &3.5 \\ 
CharNet~\cite{xing2019convolutional} &92.7 &90.5 & 91.6 &- \\
TextFuseNet~\cite{ye2020textfusenet} &\textbf{94.0}	&\textbf{90.6}	&\textbf{92.2} &4.1 \\
\hline
RSCA-640 &85.3 &81.3 &83.2 &\textbf{32.9} \\  
RSCA-800 &87.2 &82.7 &84.9 &23.3 \\

\bottomrule
\end{tabular}
\caption{Detection results on ICDAR-2015 dataset. All results are collected both from ICDAR-2015 leaderboard and original papers. Hyper-parameters of RSCA are adopted directly from CTW1500.}
\label{Table 4}
\end{table}

\subsection{Experimental Settings}
\noindent \textbf{Training and inference setting} We build the whole RSCA model illustrated in Figure~\ref{Figure 2}. At first, we pre-train all models on SynthText dataset for 2 epochs. Then, we fine-tune our models on each dataset for 1200 epochs with stochastic gradient descent (SGD). For each dataset, we set batch-size to 16 with synchronized batch normalization. For learning rate policy, we employ a poly learning rate decay in which the initial learning rate is multiplied by $(1-\frac{epoch}{max_{epoch}})^{power}$, where the initial learning rate is set to be 0.007 and $power$ is set to be 0.9. We also use a weight decay of 0.0001 and a momentum of 0.9. Data augmentation are mainly listed as follows: (1) Images are randomly horizontally flipped and rotated in the range $[-10^{\circ},10^{\circ}]$; (2) Images are randomly reshaped with ratio $[0.5, 3.0]$ and then cropped by 640 × 640 samples for training efficiency. These training setups mostly follow previous methods in DBnet~\cite{liao2020real} and PSENet~\cite{Wang_2019_CVPR} for fair comparisons and quick setting ups. All RSCA models are trained based on 4 NVIDIA V100 GPUs and tested on a single V100 GPU. The RSCA framework is implemented based on MegReader toolbox.
\\


\begin{table}[tb]
\centering
\begin{tabular}{c|c|c|c}
\toprule
Method    & Precision   &Recall  & F-measure \\ 
\midrule[1pt]
Nearest &82.7 &78.9 &80.8  \\
Bilinear &82.9 &79.0 &80.9  \\
Deconvolution &82.9 &79.5 &81.2 \\
Pixel Shuffle  &83.2 &78.8 &80.9 \\
Spatial Attention~\cite{fu2019dual} &84.2 &77.5 &80.7 \\
Channel Attention~\cite{fu2019dual} &84.9 &78.2 &81.3 \\
LCAU-FPN   &85.8 &77.9 &81.8 \\
LCAU-All &86.7 &78.8 &82.6 \\
\bottomrule
\end{tabular}
\caption{Detection results on CTW1500 dataset with different upsampling operators.}
\label{Table 5}
\end{table}

\begin{table}[tb]
\centering
\begin{tabular}{c|c|c|c|c}
\toprule
Backbone    & Precision   &Recall  & H-mean & Size(M)\\ 
\midrule[1pt]
ResNet-50 &86.7 &78.8 &82.6 &28.18 \\
ResNet-101 &86.0 &79.9 &82.8 &47.18 \\
Mobilenetv3  &81.7 &73.8 &77.5 &6.89 \\
EfficientNet-b0  &84.1 &75.7 &79.7 &6.54 \\
EfficientNet-b1  &83.8 &75.5 &79.4 &9.04 \\
EfficientNet-b2  &84.9 &75.5 &79.9 &10.37 \\
\bottomrule
\end{tabular}
\caption{Detection results on CTW1500 dataset with different backbones of RSCA. Size refers to model size including backbone and feature pyramids.}
\label{Table 6}
\end{table}

\begin{table}[tb]
\centering
\begin{tabular}{c|c|c|c}
\toprule
Loss function    & Precision   &Recall  & H-mean\\ 
\midrule[1pt]
BCE loss &82.0 &75.7 &78.7 \\
BCE loss++ &83.2 &78.7 &80.9 \\
Focal loss\cite{lin2017focal}  &83.9 &72.3 &77.6  \\
\bottomrule
\end{tabular}
\caption{Detection results on CTW1500 dataset with different loss functions for RSCA.}
\label{Table 7}
\end{table}

\begin{table}[tb]
\centering
\begin{tabular}{c|c|c}
\toprule
Feature Aggregations    & F-measure   & Model Size(M)\\ 
\midrule[1pt]
FPN-64c~\cite{lin2017feature} &81.9 &26.24 \\
BiFPN-D1-64c~\cite{tan2020efficientdet} &81.8 &26.40  \\
BiFPN-D2-88c~\cite{tan2020efficientdet} &81.1 &26.60  \\
BiFPN-D3-112c~\cite{tan2020efficientdet} &80.4 &26.88  \\ 
\bottomrule
\end{tabular}
\caption{Detection results on CTW1500 dataset with different feature aggregations for RSCA.}
\label{Table 8}
\end{table}

\begin{table}[tb]
\centering
\begin{tabular}{c|c}
\toprule
Component    & Time Cost (ms) \\ 
\midrule[1pt]
Mobilenetv3 &18.66 \\
Post-processing &10.65  \\
Total &29.31  \\
\bottomrule
\end{tabular}
\caption{Inference time of Mobilenetv3-based RSCA.}
\label{Table 9}
\end{table}

\begin{figure}[tb]
\centering
\includegraphics[width=0.5\textwidth]{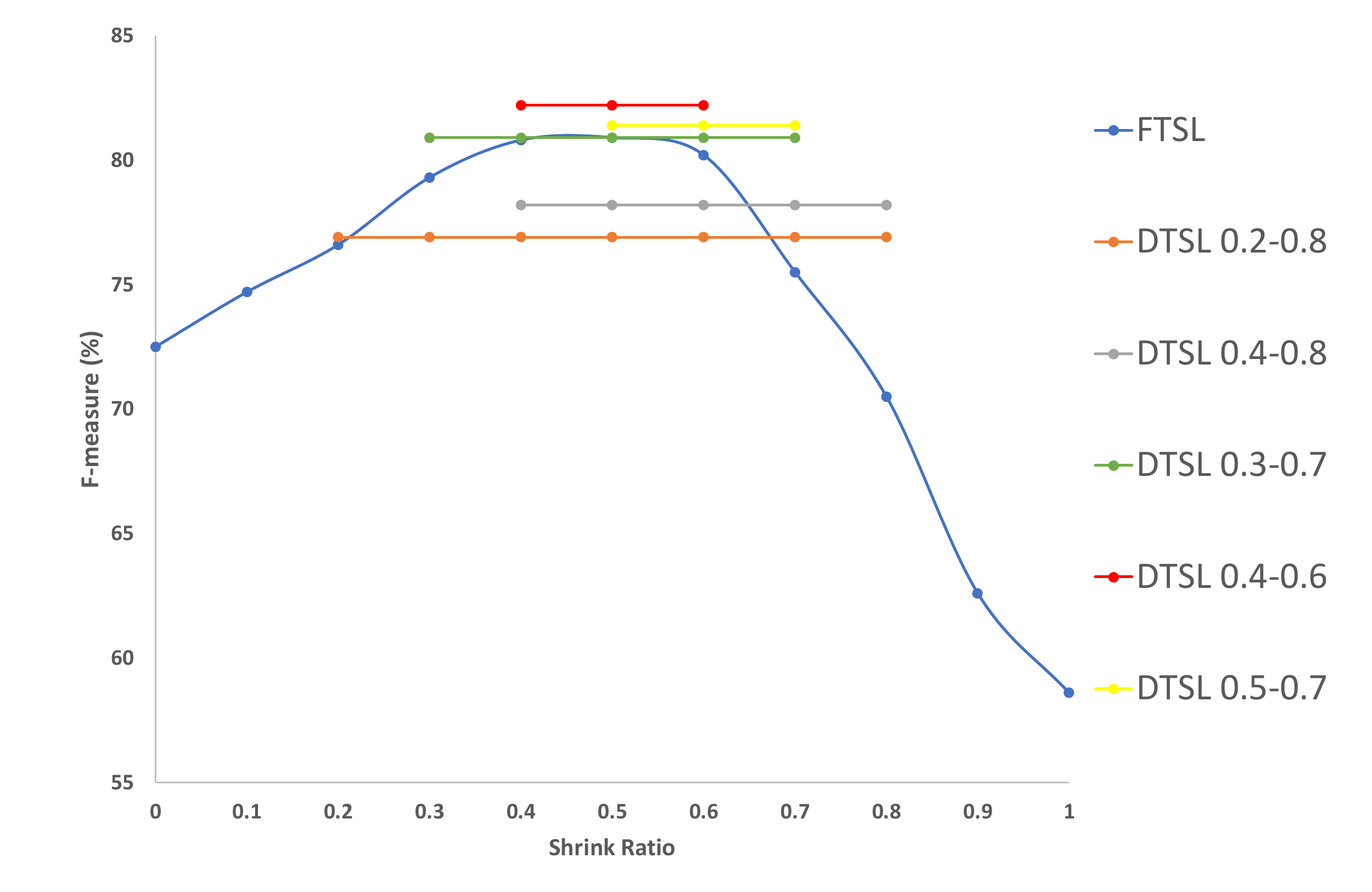}
\caption{Comparisons between dynamic and fixed text-spine labeling with RSCA on CTW1500 dataset.} 
\label{Figure 5}
\end{figure}

\subsection{Comparisons with State-of-the-arts}
We conduct extensive experiments on two curved text detection benchmarks CTW1500 and Total-Text, two multi-oriented text detection benchmarks MSRA-TD500 and ICDAR-2015. All hyper-parameters are tuned based on CTW1500 and directly employed on other datasets.
\\

\noindent \textbf{Curved Text Detection} To tackle with curved text detection, we mainly perform experiments on CTW1500 and Total-Text datasets. All experiments with CTW1500 and Total-Text are shown in Table~\ref{Table 1} and Table~\ref{Table 2}. 
\\

\noindent \textbf{Multi-Oriented Text Detection} To tackle with multi-oriented text detection, we mainly perform experiments on MSRA-TD500 and ICDAR-2015 datasets. All experimental results with MSRA-TD500 and ICDAR-2015 are shown in Table~\ref{Table 3} and Table~\ref{Table 4}. 

\begin{figure*}[htb]
\centering
\subfigure{
\centering
\includegraphics[width=0.23\textwidth,height=0.12\textheight]{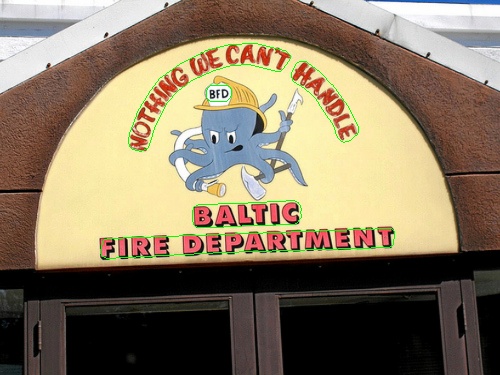}}
\subfigure{
\centering
\includegraphics[width=0.23\textwidth,height=0.12\textheight]{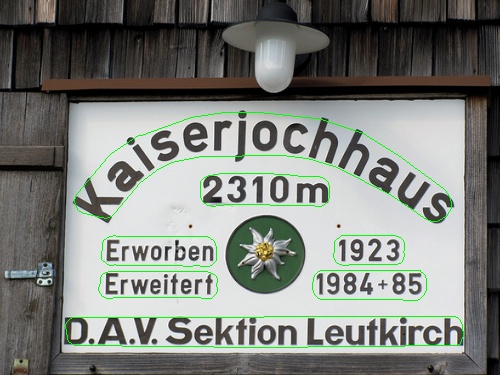}}
\subfigure{
\centering
\includegraphics[width=0.23\textwidth,height=0.12\textheight]{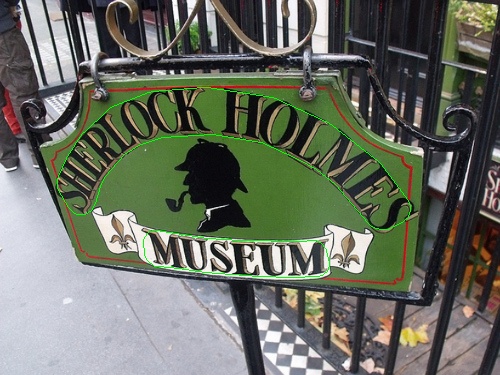}}
\subfigure{
\centering
\includegraphics[width=0.23\textwidth,height=0.12\textheight]{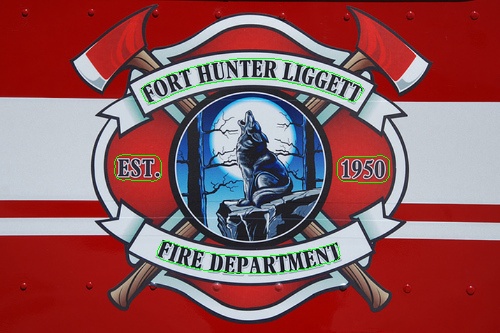}}

\subfigure{
\centering
\includegraphics[width=0.23\textwidth,height=0.12\textheight]{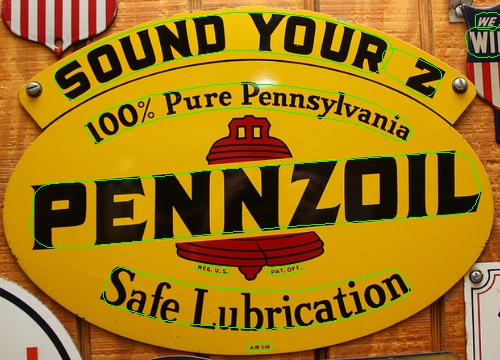}}
\subfigure{
\centering
\includegraphics[width=0.23\textwidth,height=0.12\textheight]{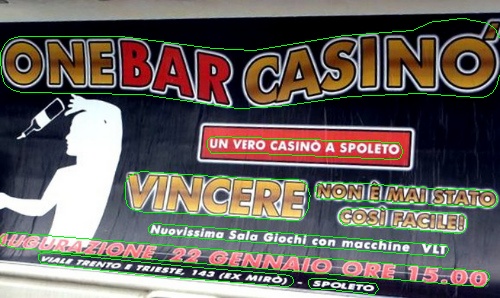}}
\subfigure{
\centering
\includegraphics[width=0.23\textwidth,height=0.12\textheight]{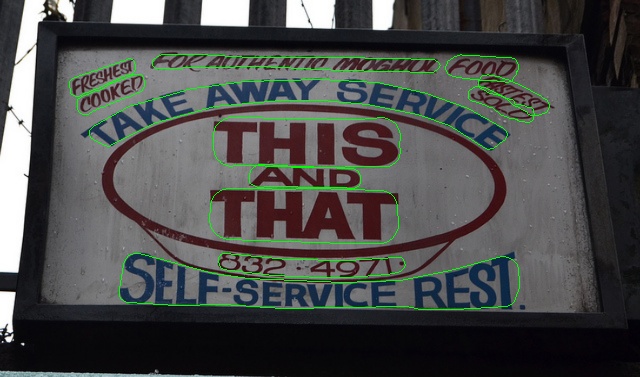}}
\subfigure{
\centering
\includegraphics[width=0.23\textwidth,height=0.12\textheight]{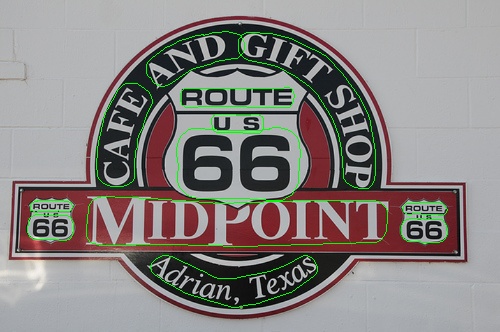}}

\subfigure{
\centering
\includegraphics[width=0.23\textwidth,height=0.12\textheight]{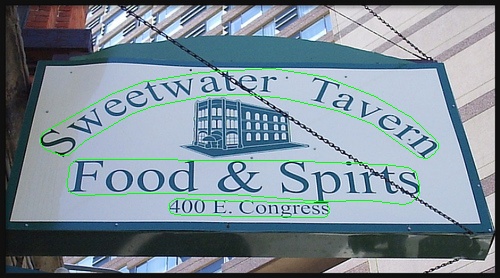}}
\subfigure{
\centering
\includegraphics[width=0.23\textwidth,height=0.12\textheight]{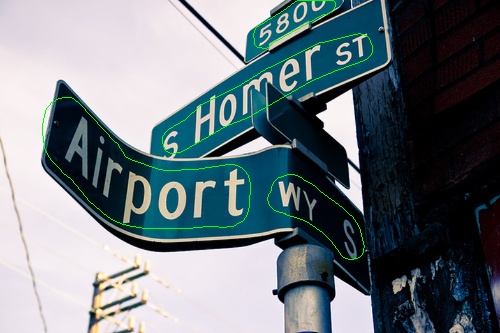}}
\subfigure{
\centering
\includegraphics[width=0.23\textwidth,height=0.12\textheight]{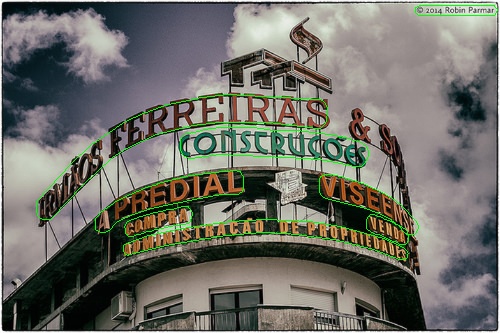}}
\subfigure{
\centering
\includegraphics[width=0.23\textwidth,height=0.12\textheight]{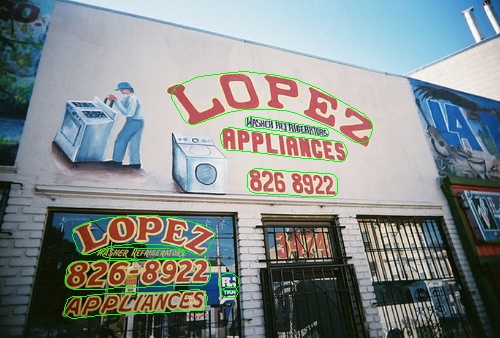}}

\caption{Visualization of detection results. Images are selected from test set of CTW1500 dataset.}
\label{Figure 6}
\end{figure*}

\subsection{Ablation Study}
\noindent \textbf{Local Context-aware Upsampling} We study the effectiveness of our local context-aware upsampling module by comparing it with different upsampling operators. We use an Imagenet~\cite{deng2009imagenet} pre-trained ResNet-50 as backbone and build the whole model in Figure~\ref{Figure 2}. In Table~\ref{Table 5}, we implement different upsampling methods and it shows that our local context-aware upsampling brings consistent improvements on precision and outperforms previous upsampling methods. 
\\

\noindent \textbf{Dynamic Text-Spine Labeling} To validate the effectiveness of dynamic text-spine labeling, we make comparisons with same models under different constant text-spine shrink ratios in Figure~\ref{Figure 5}. FTSL is fixed text-spine labeling that uses a constant shrink ratio with RSCA model, which achieves best performances around 80.5\% F-measure when shrink ratio is 0.4 or 0.5 on CTW1500 dataset. For dynamic text-spine labeling that shrinks from $r_a$ to $r_b$, it shows that the best setting with $r_a = 0.4$ and $r_b = 0.6$ could achieve 82.1\% F-measure with the same model.
\\

\noindent \textbf{Different Backbones} We evaluate our RSCA with different backbones in Table~\ref{Table 6}. It includes ResNet with different depth of layers, EfficientNet~\cite{tan2019efficientnet} with different scales and MobileNetv3~\cite{howard2019searching}. All backbones are pre-trained on ImageNet~\cite{deng2009imagenet} and it shows that our RSCA is compatible with these state-of-the-art light-weight backbones, which can be deployed on mobile devices.
\\

\noindent \textbf{Different Loss Functions} We evaluate our RSCA with different loss functions in Table~\ref{Table 7}. It includes basic binary cross entropy loss (BCE loss), binary cross entropy loss with hard negative mining (BCE loss++) and focal loss~\cite{lin2017focal}. 
It shows that binary cross entropy loss with hard negative mining outperforms others. Setting different loss functions usually require many hand-crafted fine-tuning parameters like binary thresholds and shrink ratios. The way to find the most suitable loss function for scene text detection remains an open problem for the community.
\\

\noindent \textbf{Different Feature Aggregations} We evaluate our RSCA with different feature aggregations in Table~\ref{Table 8}. We mainly implement basic FPN~\cite{lin2017feature} and BiFPN~\cite{tan2020efficientdet} with different repeated times. FPN-64c means the channels of all feature maps are 64 during the feature aggregation stage and the same to BiPFN. It shows that repeated feature aggregations with more scalings are not helpful for improving performance of scene text detection because the bottleneck now is the lack of modeling spatial transformation on local ranges.
\\

\noindent \textbf{Mobile Device Inference} To analyze the performance of our model on mobile devices, we use the RSCA model based on Mobilenetv3~\cite{howard2019searching} backbone, which can achieve 34.11 FPS on a single V100 GPU and shown in Table~\ref{Table 9}. Since the inference time of Mobilenetv3 is 192ms on a Snapdragon 660 CPU, according to the AI-Benchmark~\cite{ignatov2018ai}. An corresponding estimation of inference time of RSCA model on a Snapdragon 660 CPU based mobile phone like Redmi Note 7, would be around 302 ms, which is around 3 FPS. For a more powerful mobile CPU like Snapdragon 855, which takes 47ms for Mobilenetv3 inference according to the AI-Benchmark~\cite{ignatov2018ai}, our RSCA can reach an estimation inference time at 74ms, which is around 13.5 FPS. More accelerations and optimizations like model pruning and quantization can be further employed for mobile device deployment.

\subsection{Visualization}
We visualize some detection results of our RSCA model on test set of CTW1500 dataset in Figure~\ref{Figure 6}. It shows that our model can accurately detect most of curved text instances in different scenes, including text on billboards, traffic signs and slogans.

\section{Conclusion}
In this paper, we first analyze the problems and bottlenecks of current segmentation-based models for arbitrary-shaped scene text detection, mainly on the lack of geometrical modeling and complex label assignments or repeated feature aggregations. To tackle these problems, we propose RSCA: a Real-time Segmentation-based Context-Aware model for arbitrary-shaped scene text detection with local context-aware upsampling and dynamic text-spine labeling, which models local spatial transformation and simplifies label assignments separately. Our experiments show that RSCA achieves state-of-the-art performances with real-time inference speed on multiple arbitrary-shaped scene text detection benchmarks.

{\small
\bibliographystyle{ieee_fullname}
\bibliography{egbib}
}

\end{document}